\documentclass[journal,onecolumn,12pt,draftclsnofoot]{IEEEtran}
\IEEEoverridecommandlockouts

\onecolumn

\usepackage{cite}
\usepackage{amsmath,amssymb,amsfonts}
\usepackage{algorithmic}
\usepackage{graphicx}
\usepackage{textcomp}
\usepackage{hyperref}
\usepackage{tikz,pgfplots}
\pgfplotsset{width=10cm,compat=1.9}
\usetikzlibrary{calc}
\usepackage{algorithm}

\usepackage{pgfplotstable}
\usetikzlibrary{positioning}
\usetikzlibrary{pgfplots.groupplots}
\usetikzlibrary{arrows}

\usepackage{xcolor,mathcomSTEv4}
\def\BibTeX{{\rm B\kern-.05em{\sc i\kern-.025em b}\kern-.08em
    T\kern-.1667em\lower.7ex\hbox{E}\kern-.125emX}}

\newcommand{\topK}{ {\rm top} K}

\newcommand{\comp}{ {\rm comp}}

\usepackage{amsthm}
\newtheorem{rem}{Remark}

\begin{document}

\title{
M22: A Communication-Efficient Algorithm for Federated Learning Inspired by Rate-Distortion
 %
}

\author{Yangyi Liu,
        Stefano Rini,
        Sadaf Salehkalaibar,
        and Jun Chen
\thanks{Y. Liu and J. Chen are with the Department
of Electrical and Computer Engineering, McMaster University, Hamilton,
ON, L8S 4L8 Canada, e-mails: liu5@mcmaster.ca and chenjun@mcmaster.ca}
\thanks{S. Rini is with National Yang-Ming Chiao-Tung University (NYCU), e-mail: stefano.rini@nycu.edu.tw}
\thanks{S. Salehkalaibar is with University of Toronto, e-mail: sadaf.salehkalaibar@utoronto.ca}}

\maketitle
\vspace*{-10mm}
 
\begin{abstract}

In federated learning (FL),  the communication constraint between the remote learners and the Parameter Server (PS) is a crucial bottleneck. 
For this reason, model updates must be compressed so as to minimize the loss in accuracy resulting from the communication constraint.
This paper  proposes ``\emph{${\bf M}$-magnitude weighted $L_{\bf 2}$ distortion + $\bf 2$ degrees of freedom''} (M22) algorithm, a rate-distortion inspired approach to gradient  compression for federated training of deep neural networks (DNNs).

In particular, we propose a  family of distortion measures between the original gradient and the reconstruction we referred to as ``$M$-magnitude weighted $L_2$'' distortion, and we assume that gradient updates follow an i.i.d. distribution -- generalized normal  or Weibull, which have two degrees of freedom. 
In both the distortion measure and the gradient, there is one free parameter for each that can be fitted as a function of the iteration number.
%
Given a choice of gradient distribution and distortion measure, we design the quantizer minimizing the expected distortion in gradient reconstruction. 
To measure the gradient compression performance under a communication constraint, we define the \emph{per-bit accuracy} as the optimal improvement in accuracy that one bit of communication brings to the centralized model over the training period. 
Using this performance measure, we systematically benchmark the choice of gradient distribution and distortion measure.
We provide substantial insights on the role of these choices and argue that significant performance improvements can be attained using such a rate-distortion inspired compressor.

\end{abstract}


\section{Introduction}
Federated learning (FL) holds the promise of enabling the distributed training of large models over massive datasets while preserving data locality, guarantying scalability, and also preserving data privacy.
Despite the great advantages promised by FL, the communication overhead of distributed training poses a challenge to contemporary networks.
As the size of the trained models  and the number of devices participating to the training is ever increasing, the transmission from remote users to the parameter server (PS) orchestrating the training process becomes the critical performance bottleneck  \cite{konevcny2016federated,li2019federated}.
In order to address this issue, the design of an effective gradient compression algorithm
is of paramount importance. 
In this paper, we propose M22 -- a gradient compression algorithm inspired by rate-distortion principles. More specifically, M22 performs gradient quantization using (i) a family of distortion metrics which provides higher precision for higher at gradient magnitudes and (ii) under the assumption that the gradients as i.i.d. samples from a distribution.
We show that, by leveraging these two modelling choices, one can design an efficient gradient compressor in FL scenarios. 
More specifically,  we measure the compression performance in terms of the \emph{per-bit accuracy}, that is the largest improvement in accuracy that can be attained, on average, by one-bit communication between the remote users and the PS. 
Through this performance measure, we are able to show that the superiority of M22 over other approaches in the literature.

%
%
%
%
 %
 \subsection{Literature Review}
In recent years, distributed learning has received considerable attention in the literature \cite{bertsekas2015parallel}. 
In the following, we shall briefly review the contributions dealing with communication aspects of FL  which are most relevant to the development of the paper. 

From a general perspective, FL consists of a central model which is trained  locally at the remote clients by applying Stochastic Gradient Descent (SGD) over a dataset present at the users.
The local gradients are then communicated to the central PS for aggregation into a global model.
Since this aggregation model does not require data centralization, it provides substantial advantages in terms of scalability, robustness, and security \cite{li2020federated}. 
For these reasons, there has been a significant interest in developing effective and efficient FL algorithms  \cite{Shalev-Shwartz2010FL_CE,Wang2018Spars_FL,Alistarh2018Spars_FL}.
%

When considering a concrete deployment of FL algorithms, the performance bottleneck is often found in the communication rates between the remote users and the PS.  
This scenario is often referred to as the \emph{rate-limited FL} scenario \cite{shlezinger2020communication}.
In this case, one is interested in determining the relationship between the accuracy and the uplink communication rate \cite{shlezinger2020communication,saha2021decentralized}.
%
%
For the rate-limited FL settings, researchers have focused on the problem of  designing  algorithms that efficiently  compress the local gradient at the remote users \cite{SGD_sparse2}.
Generally speaking, gradient compression algorithms can be divided in two classes: gradient sparsification \cite{Shalev-Shwartz2010FL_CE,Wang2018Spars_FL,Alistarh2018Spars_FL}
and (ii) gradient quantization  \cite{seide2014onebitSGD,Konecny2016Fl_CE, gandikota2019vqsgd,Bernstein2018signSGC,FL_DSGD_binomial,Li2019DP_CEFL}. 

A complementary approach  consists in applying some dimensionality reduction algorithms to the gradients before compression.
For instance, the authors of  \cite{NEURIPS2019_75da5036} consider a scheme in which each client performs local compression to the local stochastic gradient by count sketch via a common sketching operator. 
When gradient dimensionality reduction and/or compression are employed, the training performance can be improved through error feedback \cite{stich2018sparsified,rothchild2020fetchsgd}. 
%
In other literatures, the gradient compression is performed on the whole gradient vector, rather than entry-wise.  For instance,  the authors of \cite{gandikota2019vqsgd} introduce vector  quantization for SGD.

 More naively,  constraints in the communication capabilities 
 between the remote users and the PS can also be addressed by restricting the number of communication iterations between gradient updates  \cite{McMahan2018DP_CEFL,hu2020cpfed}. 
%
From an implementation-oriented perspective, \cite{sun2019hybrid} studies the effect of gradient quantization when constrained to a \emph{sign-exponent-mantissa} representation.

The design of good gradient quantizers sometimes relies on the assumption on the gradient distribution.
Assuming that DNN gradient entries are i.i.d. distributed according to some distributions is a powerful approximation which is adopted in various contexts, from network pruning to inference modelling. %
Some authors assume that gradients have i.i.d Gaussian \cite{khan2019approximate} or Laplace  \cite{isik2021successive} entries.
Other authors consider distributions with two degrees of freedom, such as generalized Normal \cite{chen2021dnn} or two-sided Weibull  \cite{fu2020don}.

 \subsection{Contributions}

In this paper we consider the federated training, where the communication constraint is applied on the transmission from the remote clients to the PS, while the transmission from the PS to the remote clients is unconstrained. 
In this scenario, we propose M22,  a novel gradient compression algorithm inspired by rate-distortion principles. 
More specifically, M22  relies on \underline{two} rate-distortion principles to design a scalar quantizer that meets the communication  constraint: 

\smallskip
\noindent
$\bullet$ \underline{{\bf M2}}: \emph{${\bf M}$-magnitude weighted $L_{\bf 2}$}  distortion measure-- a choice of a distortion measure between the original and the compressed gradient: the ``$M$-magnitude weighted $L_2$'' distortion. This distortion promotes higher fidelity for higher gradient entries and reflects the practitioner's intuition that larger gradients have larger impact on the model updates.

\smallskip
\noindent
$\bullet$ \underline{{\bf 2}}: \emph{$\bf 2$ degrees of freedom} distribution fitting -- the fitting of the gradient distribution using a distribution having two degrees of freedom:   the generalized normal distribution (GenNorm) distribution \cite{chen2021dnn}, or the two-sided Weibull distribution \cite{fu2020don}.
Such choice of distributions allows one to match the variance of the gradient sample distribution, as well as the tail decay as it evolves through the iteration number.

%
%
%

To measure the compression performance in the rate-limited FL setting, we introduce the concept of \emph{per-bit accuracy} as the relevant performance measure for distributed training under communication constraints.
The per-bit accuracy corresponds to the  improvement in accuracy that a gradient compressed within $\Rsf$ bits can provide to a given model.
Using this performance measure, we deploy extensive empirical evaluations  of the performance of M22 for different DNN networks, communication rates, and choice of hyperparameters. 
%
%
%
Also, we provide extensive  numerical simulation results to show that  M22 outperforms other approaches in the literature.
\medskip
\noindent
{\bf Notation:} In the following, lower case boldface letters (eg. $\zv$) are used for column vectors and uppercase boldface letters (e.g., $\Mv$) designate matrices. 
%
%
The all-zero vector of dimension $d$ is indicated as $\zerov_d$.
We also adopt the shorthands $[m:n] \triangleq \{m, \ldots, n\}$
and  $[n] \triangleq \{1, \ldots, n\}$. 
The $p$-norm of the vector $\xv$ is indicated as $\| \xv\|_p$.
%
%
%
Calligraphic scripts are used to denote sets (e.g., $\Acal$) and $|\Acal|$ is used to denote its cardinality.

The code for the numerical evaluations of this paper is provided online at \url{https://github.com/yangyiliu21/FL\_RD}.


 
\section{System Model}
\label{sec:System Model}
In the following, we consider the distributed training of a machine learning (ML) model across $N$ devices where the communication between the remote device and the PS is limited to $R$ bits per learner. 
We introduce the per-bit accuracy as the performance measure that allows for the comparison across gradient compression algorihtms. 
Finally, the problem is specialized to the federated DNN training scenario. 

%
%

\subsection{Distributed Optimization Setting}
\label{sec:Optimization Setting}

Consider the setting with $N$ clients each possessing a local dataset $\Dcal_n=\{ \dv_{nk} \}_{k \in [|\Dcal_n|] }$ for $n \in [N]$ and wishing to minimize the \emph{loss function} $\Lcal$ as evaluated across all the local datasets     and over the choice of model $\wv \in \Rbb^d$, that is 
\ea{
\Lcal(\wv) =  \f 1 {\sum_{n \in [N]} |\Dcal_n|} \sum_{n \in [N]} \sum_{\dv_{nk} \in [\Dcal_n]} \Lcal(\dv_{nk}, \wv).
\label{eq:loss}
}
%
For the  {loss function} in \eqref{eq:loss}, we assume that there exists a unique minimizer $\wv^*$ of \eqref{eq:loss}, that is, 
\ea{
\wv^*= \argmin_{\wv \in \Rbb^d} \ \ \Lcal(\wv).
\label{eq:minimum w}
}
A common approach for numerically determining the optimal value in \eqref{eq:minimum w} in the centralized scenario is through the iterative application of (synchronous) stochastic gradient descent (SGD).
In the (centralized) SGD algorithm, the learner maintains an estimate of the minimizer in \eqref{eq:minimum w}, $\wv_t$, for each time $t\in[T]$. The final estimate of \eqref{eq:minimum w} is $\wv_T$. 
At each time $t\in[T]$, the estimate $\wv_t$ is updated as
%
%
\ea{
\wv_{t+1}=\wv_{t}-\eta_t  \gv_t,
\label{eq:SGD}
}
for $\wv_{0}=\zerov_d$, where $\eta_t$ is  an iteration-dependent step size $\eta_t$ called \emph{learning rate}, and where  $\gv_t$ is the stochastic gradient of $\Lcal$ evaluated at $\wv_{t}$, that is 
\ea{
\Ebb\lsb \gv_t\rsb= \sum_{ \dv_k \in \Dcal } \nabla\Lcal\lb \dv_k, \wv_t\rb.
\label{eq:mean gradient}
}
In \eqref{eq:mean gradient}, $\nabla\Lcal\lb\wv_n\rb$ denotes the gradient of $\Lcal\lb \dv_k,\wv_t\rb$ at $\wv_t$ as evaluated over the dataset  $\Dcal = \bigcup_{n \in [N]} \Dcal_n$.
%
%
%
%
%

In the FL setting, given that the datasets $\Dcal_{n}$ are distributed at multiple remote learners, the SGD algorthim as in \eqref{eq:SGD}  has to be adapted as follows. 
%
%
%
%
First (i) the PS  transmits the current model estimate, $\wv_t$,
to each client $n \in[N]$,
%
%
then (ii) each  client $n \in [N]$ accesses its  local dataset $\Dcal_{n}=\left\{\lb \dv_{n}(k),v_{n}(k) \rb\right\}_{k\in\left[\left|\Dc_{n}\right|\right]}$ and computes the stochastic gradient, $\gv_{nt}$, as in \eqref{eq:mean gradient} and communicates it to the PS.
Finally (iii) the PS updates the model estimate as  in \eqref{eq:SGD} but where $\gv_t$ is obtained as
\ea{
\gv_t=\f{1}{N}\sum_{n \in[N]}\gv_{nt}.
\label{eq:aggregate}
}
%
The distributed version of SGD for the FL setting is referred to as \emph{federated averaging} (FedAvg)  \cite{konevcny2016federated}.

\subsection{Federated Learning with Communication Constraints}
\label{sec:Communication Setting}
Customarily, in the FL setting, the communication is assumed to take place over some noiseless, infinity capacity link connecting the PS and the remote users and vice-versa. 
In a practical scenario, the users model  wireless mobiles, IoT devices, or sensors which have significant limitations in the available power and computational capabilities.
%
In these scenarios, we can still assume that  users rely on some physical and MAC layers' protocols that are capable of reliably delivering a certain payload from the users to the PS. 

For this reason, in the following, we assume that the communication between each of the remote users and the PS takes place over a rate-limited channel of capacity $d \Rsf$, where $d$ is the dimension of the model in Sec. \ref{sec:Optimization Setting}.
In other words, each client can communicate up to $d \Rsf$ bits for each iteration $t \in [T]$.

In the following, we refer to the operation of converting the $d$-dimensional gradient vector $\gv_{nt}$ to a $d \Rsf$ binary vector as \emph{compression}. 
%
Mathematically, compression is indicated though the operator
\ea{
\comp_{\Rsf}: \ \   \Rbb^d \goes  [2^{d\Rsf}].
\label{eq:comp}
}
Similarly, the reconstruction of the gradient is denoted by $\comp_{\Rsf}^{-1}$.
Note that in \eqref{eq:comp}, $\Rsf$ indicates the number of bits per model dimension. 

\begin{rem}
\label{rem:all the same}
For simplicity, in the following we assume  that (i) all users are subject to the same communication constraint and (ii) all users employ the same set of compressors. 
Generalizing the results in the paper to the more general scenario in which remote users have different communications constraints and use different compressor is rather straight-forward. 
%
\end{rem}

Under the assumptions in Rem. \ref{rem:all the same}, the model update in \eqref{eq:SGD} can be reformulated as 
%
\ea{
\whv_{t+1}& =\whv_{t}-\eta_t  \ghv_t \nonumber \\
\ghv_t & = \f 1 n \sum_{n \in [N]} \comp_\Rsf^{-1} \lb\comp_\Rsf (\gv_{tn}) \rb,
\label{eq:SGD comp}
}
with  $\whv_0 = \zerov$.
%
%
%
For the model update rule in \eqref{eq:SGD comp}, it is possible to derive similar convergence guarantees to that for the unconstrained problem in \eqref{eq:SGD} -- see \cite{SGD_sparse2}.

\subsection{Compression Performance Evaluation}
\label{sec:Compression Performance Evaluation}
In the following, we are interested in characterizing the compression performance in terms of the loss of accuracy as a function of the communication rate.
More formally, given the model estimate $\whv_t$  and the gradient estimate $\ghv_t$, we wish determine 
\ea{
\Gsf_\Rsf(\whv_{t+1}) = \min_{\comp_{\Rsf},\comp_{\Rsf}^{-1}} \Lcal(\whv_{t+1}),
\label{eq:G}
}
where $\whv_{t+1}$ is obtained as in \eqref{eq:SGD comp}.
%

In general, we are interested in determining the effect of compression through the SGD iterations: to this we define 
\ea{
\Delta (T,\Rsf) =  \f 1 { d \Rsf } \f {\lb \Lcal(\wv_T) - \Gsf_{\Rsf}(\whv_{T}) \rb} T
\label{eq:pba}
}
as the \emph{per-bit accuracy}. 
This definition \eqref{eq:pba} corresponds to the overall loss of accuracy due to compression of the gradient to $\Rsf$-bits per dimension at the training horizon, $T$.
By comparing   $\Delta (T,\Rsf)$ for different values of $T$ and $\Rsf$ on the same optimization problem and a number of remote users, one can gauge the impact of the communication constraint over the training process at hand. 

Generally speaking, the minimization in \eqref{eq:G} is  too complex, as the loss function $\Lcal$ is generally non-convex in the model $\wv$. 
Additionally, lacking  a statistical description of the SGD  process is available, it is impossible to resort to classical compression techniques from information theory. 
To address these difficulties, later in Sec.  \ref{sec:A Rate-distortion Approach}, we consider a rate-distortion approach in which we simplify  the minimization in \eqref{eq:G} for the case of DNN training by considering a family of  distortion measures which captures the loss of accuracy as a function of the gradient magnitude, and assume that the DNN gradients are iid  draws from the GenNorm distribution or Weibull distribution. 
These two simplifications yield a compressor design which shows improved performance over other compressors considered in the literature.



\subsection{DNN Training}
\label{sec:DNN training}
While the problem formulation in Sec. \ref{sec:Compression Performance Evaluation} is rather general, in the remainder of the paper, we shall only consider the scenario of DNN training. 
%
%
More specifically, we  consider
a simple convolutional neural network (CNN)
  and two widely-used  architectures -- ResNet18 and VGG16.
 The three networks above are trained for  image classification and other computer vision tasks over the CIFAR-10 dataset. 
In Table \ref{tab:model_layers}, we list the parameter information of our three models: CNN, ResNet18, and VGG16.
During our training, the CNN model in is trained using SGD with  learning rate  $0.0001$ and cross-entropy loss. 
The ResNet18 and VGG16  models are trained using Adam with learning rate $0.001$ 
and $0.00005$, respectively. 
Other training hyper-parameters, including mini-batch size, loss function and etc. could be found in Table \ref{tab:DNN parameters}. 

%
%

In our FL setting, we randomly split the  CIFAR-10 training set and allocate to two remote clients. The distributions of two local datasets are the same. 
Additionally, the server requires the remote clients to report their local updates every time once it finishes one local training epoch. This combination of hyper-parameters ensures that the training framework fits into the FL scheme, and facilitates the convergence of the global model.

\begin{rem}
There are  numerous FL training settings of practical relevance in which the  choice of hyper-parameters is drastically different from the one we consider here.
%
%
We believe that the three networks above represent a simple and yet meaningful benchmark for the proposed approach, M22.   
%
We leave the testing on more sophisticated ones for future work. 
%
%
\end{rem}

\begin{table}[t]
\vspace{0.5cm}
\footnotesize
    \centering
        \caption{A summary of the parameter information of the models in Sec. \ref{sec:DNN training}. }
    \label{tab:model_layers}
\begin{tabular}{|c|c|c|c|c|}
\hline
Architectures      & Layers    & Total Params   & conv\_layer  & dense\_layer   \\ \hline
CNN                & 44     & 552,874   & 549,280  & 0  \\ \hline
ResNet18           & 98     & 11,184,068   & 11,171,008  & 0  \\ \hline
VGG16              & 32     & 33,638,218   & 14,714,688  & 18,882,560  \\ \hline
\end{tabular}
\vspace{-0.5cm}
\end{table}

{

\begin{table}
	\footnotesize
	\centering
	\vspace{0.04in}\caption{Parameters and hyperparameters used for the training of the DNN models.}
	\label{tab:DNN parameters}
	\begin{tabular}{|c|c|c|c|}
		\hline
        Model & CNN & ResNet18 & VGG16 \\ \hline
		Dataset & CIFAR-10 & CIFAR-10 & CIFAR-10 \\ \hline
        Training Samples & {50000} & {50000} & {50000} \\ \hline
        Test Samples & {10000}& {10000}& {10000} \\\hline
        Optimizer & SGD & Adam & Adam \\ \hline
        Learning Rate & {0.01} & {0.001} & {0.0005} \\ \hline
        Momentum & 0 & 0 & 0 \\ \hline
        Loss & Categorical Cross Entropy & Categorical Cross Entropy & Categorical Cross Entropy \\ \hline
        Epochs & 150 & 150 & 150 \\ \hline
        Mini-Batch Sizes & 64 & 64 & 32  \\ \hline
	\end{tabular}
\end{table}
}

\vspace{3cm}

\subsection{Further Comments}
%
%

Before delving further in the paper, let us clarify what aspects of the problem setting of Sec. \ref{sec:Communication Setting} will not be considered in the remainder of the paper.

\smallskip
\noindent
$\bullet$ {\bf Lossless universal compression.}
In the following, we assume that gradients are transmitted after compression. 
Note that one could further apply some lossless universal compression algorithm to further reduce the communication load. 
Such algorithm are readily available and very efficiently exploit the redundancy in the data to further reduce the transmission payload \cite{nelson1995data}. 
For simplicity, we do not consider this further compression opportunity as we it will complicate the evaluation of the effective dimension of the compressed gradient.

\smallskip
\noindent
$\bullet$ {\bf Varying the number of remote client.}
In the remainder of the paper, we consider the case of two remote clients and do not consider the effect of client scheduling (having a varying number of remote users). %
As the number of remote user varies, the overall compression error increases and thus the choice of learning rate and mini-batch size need to be adjusted accordingly. 
Although providing insight on this aspect of the hyper-parameter choice is valuable, this is outside the scope of the paper.

\smallskip
\noindent
$\bullet$ {\bf Training scheduling.}
In our setting, we consider the case in which one round of SGD is performed for each communication between the remote user and the PS, followed by a model update from the PS to the remote user. 
In general, one might consider the case in which multiple SGD rounds are performed before a model update is transmitted between the remote user and the PS.
Although this strategy also has the effect of reducing the overall communication rate, we will not consider it here.

\smallskip
\noindent
$\bullet$ {\bf Layer-wise dependency.}
%
In the following, we compress the gradients by assuming that they are independent across iterations.
Although dependencies of the gradients across layers exists, we do not consider such dependencies. 
%
For instance, in \cite{araujo2019mean},  it is shown that the per-layer distribution is conditionally dependent only on the weights in the previous layer. 
For simplicity, in the following, we do not consider this dependency. 
The design of a version of the proposed approach taking advantage of this correlation is left for future research.


\section{A Rate-distortion Approach to DNN Gradient Compression}
\label{sec:A Rate-distortion Approach}

In this section, we introduce the main ingredients of the proposed approach M22.
These ingredients further clarify the rate-distortion principles employed in  the design of the optimal compressor for DNN gradients. 
%

%
%
%
Since \eqref{eq:G} maximizing the per-bit accuracy in \eqref{eq:pba} 
as in Sec. \ref{sec:Compression Performance Evaluation} 
is generally intractable, we instead simplify the problem as follows. 
We choose (i) a  distribution to approximate the gradient entries, and (ii) distortion  that correlates the loss in accuracy in \eqref{eq:G}  when compressing the  original weights.
Once these two elements have been selected -- that a gradient distribution and a gradient distortion measure -- the compressor in \eqref{eq:comp}  is chosen as the  the quantizer which minimizes the chosen distortion for the given gradient distribution, as in the classic \cite{linde1980algorithm}.

%
%

More details about the choice of gradient distribution, distortion measure, and quantizer design are discussed in this section. 

\subsection{Gradient Distribution}
\label{sec:Gradient distribution}

%
Let us assume that gradient entries are  well-approximated as i.i.d. random variables from a certain distribution $P_{G,t}$ which varies with $t \in [T]$.  
{
We notice that mean field theory has provided a partial validation of this assumption.
In a series of papers
\cite{mei2018mean}\cite{araujo2019mean}\cite{ nguyen2020rigorous}, it has been shown that DNN weights in a given layer become  indistinguishable as the number of SGD steps grow large in various regimes.
In practice, meaningful statistics can be obtained from the gradient realization, and this justifies the adoption of this assumption from a practical perspective. 

Some efforts have been made to characterize the gradient distribution such as Laplace distribution  \cite{isik2021successive} and Gaussian distribution\cite{lee2017deep,matthews2018gaussian}. 
We refer to these distributions as one-parameter distribution due to the fact the mean of the gradients are universally assumed to be zero, leaving the only degree of freedom to be the scale of the distribution.
Consequently, we believe that they do not provide sufficient modelling capability to approximate the gradient distribution as it evolves through the iteration process. 
For this reason, we follow two approaches that approximate the gradients using either the generalized normal distribution (GenNorm) distribution in \cite{chen2021dnn}, or using a two-sided Weibull (double-Weibull or d-Weibull) distribution in \cite{fu2020don}.

Comparing with the conventional one-parameter distributions, in addition to satisfying the symmetric property, both GenNorm and d-Weibull distributions have an extra degree of freedom: the shape parameter, which substantially strengthens the layer-wise approximation capability of gradients in different models throughout the training iterations. 
Note that the generalized normal distribution encompasses the Laplace distribution when the shape parameter $\beta=1$ and normal distribution when the shape parameter $\beta=2$ as special cases. The power density function (PDF) of the GenNorm distribution is described as:
\ea{
f(x, \mu, s, \beta) = \frac{\beta}{2 s \Gamma (1/\beta)} \: e^{-(|x- \mu|/ s)^\beta}, 
\label{eq:gennorm}
}
where $\mu$ and $s$ indicate the mean and scale respectively, $\beta > 0$, and $\Gamma$ represents the gamma function: one commonly used extension of the factorial function to complex numbers. 
When the shape parameter  $1 < \beta < 2$, the distribution is leptokurtic and has fatter tail than the normal distribution. Besides the tails, the shape parameter also controls the peakedness, as it converges point-wise to a uniform distribution on $(\mu-\alpha, \mu+\alpha)$ when $\beta \xrightarrow {}\infty$. 
Despite the fact that the unimodal and symmetric properties of the GenNorm distribution coincide with the nature of DNN gradient values \cite{Bernstein2018signSGC}, Fu et al. \cite{fu2020don} proposed to approximate the gradients with a d-Weibull distribution because of its central tendency and long-tails characteristics. The PDF of the d-Weibull distribution is described as:
\ea{
f(x, \mu, s, c) = \frac{c}{2 s} \: {\left(\frac{|x-\mu|}{s}\right)}^{(c-1)} \: e^{-(|x- \mu|/ s)^c}, 
}
where $c$ is the shape parameter, and the restriction $c \in (0,1]$ guarantees the monotony of Weibull family distribution. Comparing with the Gaussian distribution and Laplace distribution, d-Weibull distribution allows the approximated distributions to be more centralized and longer-tailed, same as the GenNorm distribution. This effect corresponds to the empirical evaluation on tested models by introducing the shape parameter. An example of the fitting is provided in Fig. \ref{fig:distribution fitting}, where the bottom panel represents the simulation when more aggressive top-k sparsification is applied comparing to the top panel. 
From an empirical validation perspective, the GenNorm distribution fits the gradient histogram better than the one-parameter distributions when most of the gradients are preserved, while the d-Weibull is more suitable as the approximation distribution when more aggressive sparsification is applied.

\begin{figure}
\vspace{-0.9cm}
  \centering
  \begin{tabular}{@{}c@{}}
    \includegraphics[width=1\linewidth]{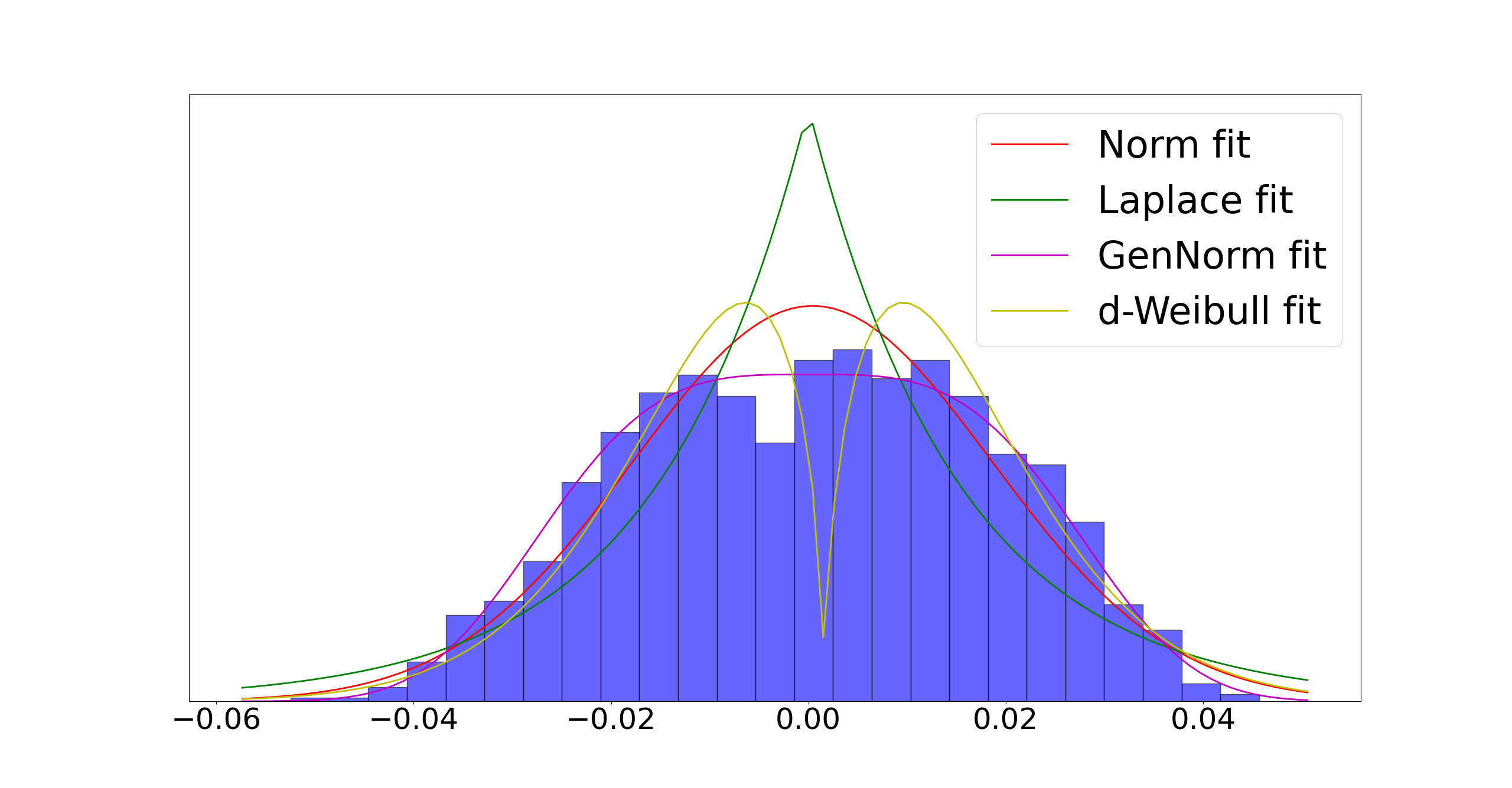}
    \vspace{-0.65cm}
  \end{tabular}
  \begin{tabular}{@{}c@{}}
    \includegraphics[width=1\linewidth]{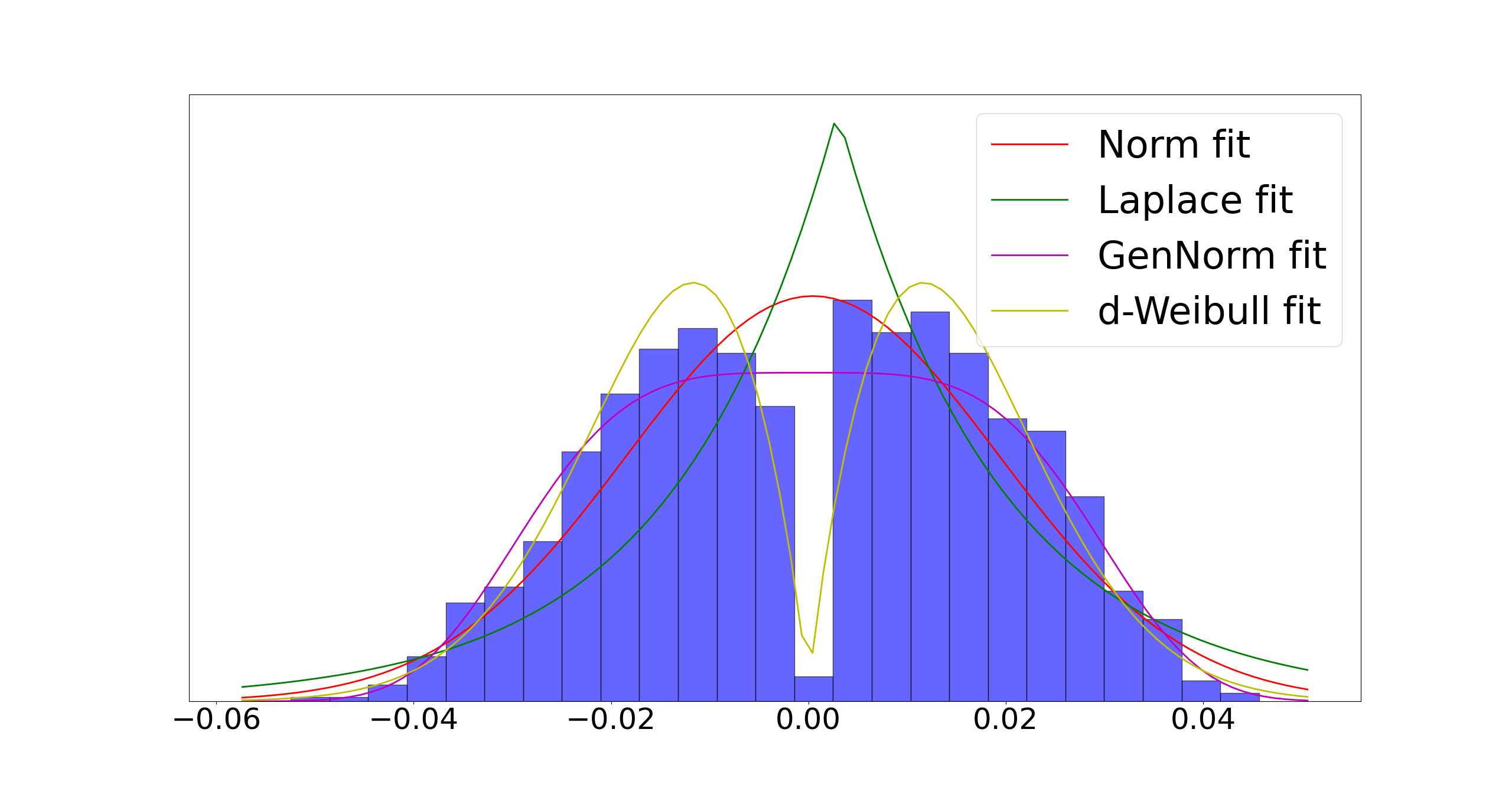} 
  \end{tabular}

  \caption{Examples of the distribution fitting of the gradient distribution as discussed in Sec. \ref{sec:Gradient distribution} for CNN, layer $42$, iteration {$10$}.}
\label{fig:distribution fitting}
\end{figure}

To better illustrate the fitting performance, let us consider Fig. \ref{fig:distribution fitting}: here we plot the gradient distribution after $\topK$ sparsification for two sparsification levels, $90\%$ and $40\%$ for the top and bottom panel respectively. 
We point out that these two percentages are chosen to demonstrate the effectiveness of GenNorm and d-Weibull fitting from a empirical validation perspective. The percentage used in practice could be other fixed numerical numbers or even dynamic. 
Generally speaking, for low sparsification levels, the generalized normal provides a tight fitting for the slow tail decay, while for high sparsification levels, the d-Weibull distribution  better matches the bi-modal nature of the empirical gradient distribution.

\subsection{Distortion Measure}
\label{sec:Distortion Measure}

The second ingredient of M22 is a judiciously chosen distortion measure between the original gradient and its reconstruction. 
We believe that this ingredient is indeed the most innovative contribution of  the paper. 
Note that the existing works mostly adopt $L_2$ loss when measuring the distortion between the original gradients and perturbed gradients.
Some exceptions \cite{isik2021successive} consider an $L_1$ distortion measure in the context of network pruning. 

In devising a distortion metric for our quantizer design, two aspects are taken into consideration: (i) the practitioner perspective on effective  sparsification  and (ii) existing bounds on the accuracy loss from gradients perturbation.
In the literature, $\topK$ sparsification  consists in setting all but the $K$ largest weights of the gradient entries to zero at each  iteration. 
It is well known among the ML practitioners that $\topK$ sparsification can be effectively used to reduce the dimensionality of the gradient updates while resulting in only a small loss in accuracy 
\cite{alistarh2018convergence,SGD_sparse2,SGD_sparse3}.
Another approach for gradient compression is uniform quantization with non-subtractive dithering \cite{QSGD,SGD_q2,SGD_q4,gandikota2019vqsgd}.
This approach finds its theoretical foundations in works such as \cite[eq. (8)]{lee2020layer} which provides a bound on the loss in accuracy as a function of the $L_2$ perturbation of the DNN weights. 

Using $L_2$ loss under extreme communication constraint circumstances performs exceptionally well. However, we found that the model convergence speed slows down when the constraint is relaxed using the same $L_2$ loss. Some intuitions behind this phenomenon could be gained as when the constraint is relaxed, we should penalize the loss harder in order to speed up the converging process. Therefore, in addition to the conventional $L_2$ term, we equip the loss with another term with a hyper-parameter that could adjust the penalizing intensity of perturbed gradients. Details and the function of this term revealed by the simulation results are discussed in \ref{sec:simu}.

Given the intuition of magnitude and two compression approaches in which either only the magnitude of the gradients is considered or the $L_2$ distortion of the gradients is used for quantizer design, we propose a new class of distortion measures combining these approaches.
That is, the ``$M$-magnitude weighted $L_2$'' distortion, mathematically defined as 
\ea{
d_{M-L_2}(\gv,\ghv) 
= \frac{1}{d}\sum_{j \in [d]} |g_j|^M  \|g_j-\hat{g}_j\|_2,
\label{eq:our distortion}
}
where $g_j$ and $\hat{g}_j$ denote the $j$-th elements of $\gv$ and $\ghv$ (gradient and quantized vectors), respectively.
%
%
Note that the parameter $M$ is again a hyper-parameter that can be used to tune the distortion to the particular iteration and instance of training. Similarly to the choice of gradient distribution, the choice of distortion measure provides us with a degree of freedom that can be adaptively adjusted to improve the training performance.




\subsection{Quantizer design}
\label{sec:Quantizer design}

Given our choice of gradient distribution and the gradient distortion measure, as in Sec. \ref{sec:Distortion Measure} and Sec. \ref{sec:Gradient distribution}, one can then construct a quantizer using the classic LGB algorithm \cite{linde1980algorithm}.
In M22, we actually apply sparsification before quantization as it is more computationally efficient to code the zero values using a run-length encoding.
The $K$-means/LGB algorithm for the class of distortions in \eqref{eq:our distortion} takes a surprisingly simple form.
In particular, let $c_k(i)$/$t_k(i)$ be the $i^{\rm th}$ centroid/threshold estimate at iteration $k$ in the scalar LGB algorithms, then
\eas{
c_{k+1}(i+1) & = \frac{\int_{t_{k}(i)}^{t_{k}(i+1)}g^{M+1} \: \text{pdf} \: (g)dg}{\int_{t(i)}^{t(i+1)} g^M \: \text{pdf} \: (g)dg},\label{eq:Kmeans-updatea}\\
t_{k+1}(i+1) & = \frac{c_{k}(i+1)+c_{k}(i)}{2},\label{eq:Kmeans-updateb}
}{\label{eq:Kmeans-update}}
for $i\in [1, 2^{\Rsf}]$ where $2^{\Rsf}$ is the number of quantization levels, $\text{pdf}(g)$ denotes the distribution fitted to the gradient vector, $c(.)$ and $t(.)$ represent the quantization centers and thresholds, respectively.

A plot of the change of quantization centers and thresholds regions versus the change of $M$ values modelled by GenNorm distribution is presented in Fig. \ref{fig:quantization regions} (because of the symmetry characteristic, only the positive regions are shown). 
A larger M choice results in more sparsed quantization regions, i.e., diverging from the center quantization bin, which corresponds to the objectives of M22, where the $\topK$ sparsification is responsible to model all small magnitude weights to the center bin, while quantization steps are in charge of modelling the long-tailed weights. 

\begin{figure}
    \centering
    \includegraphics[width=1\linewidth]{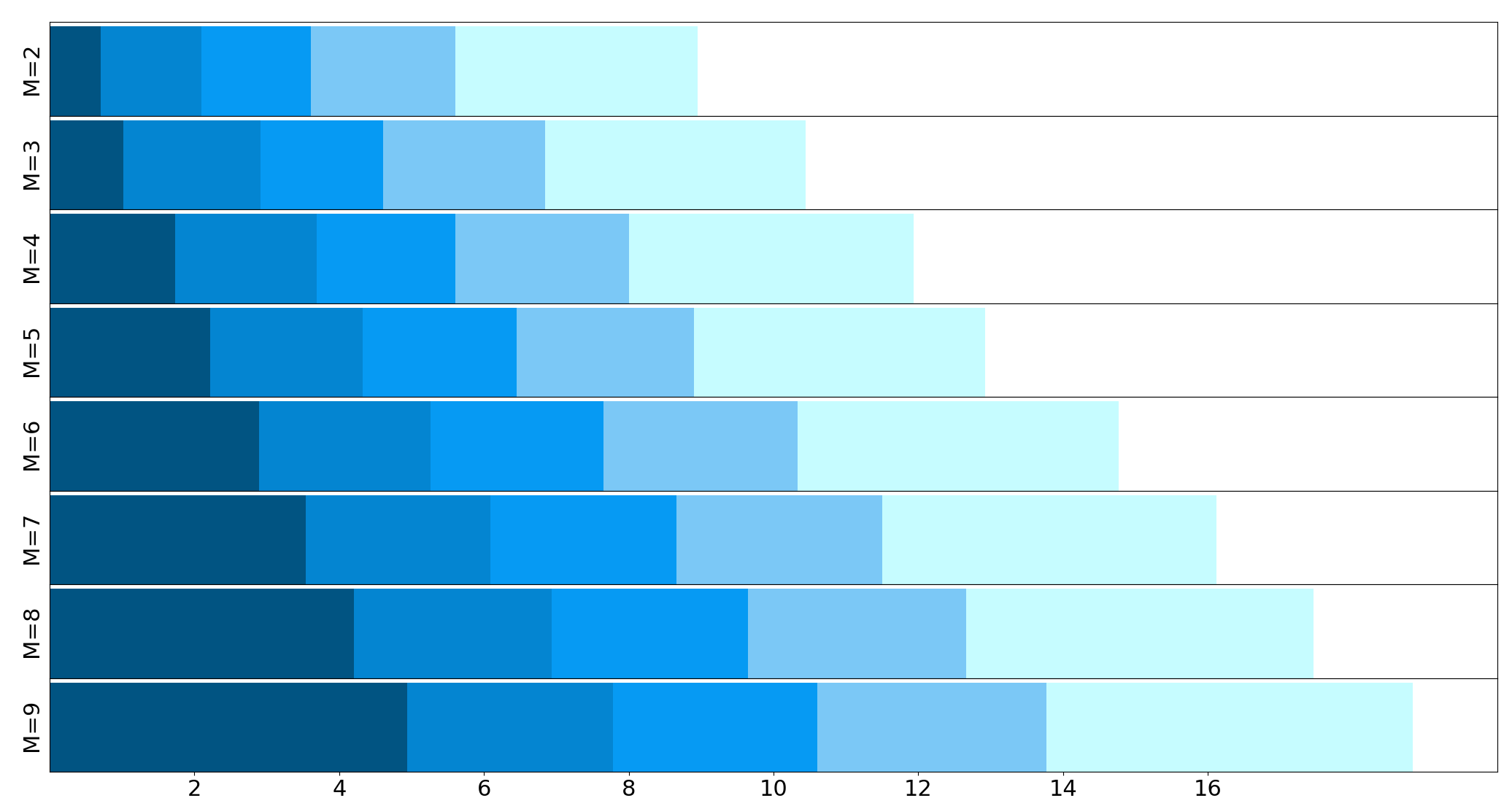}
    \caption{Quantization centers and thresholds modelled by GenNorm distribution versus M as obtained from the K-means algorithm in Sec.  \ref{sec:Quantizer design}}.
    \label{fig:quantization regions}
\end{figure}








\section{M22: the Proposed Approach}

With the demonstration of our guiding rate-distortion principles in Sec. \ref{sec:A Rate-distortion Approach},  we are finally ready to introduce M22 -- our approach inspired by rate-distortion for the design of the optimal compressor for DNN gradients. 
%

%

\subsection{Pseudo-code}
Following the famous Federated Averaging (FedAVG) algorithm proposed by \cite{pmlr-v54-mcmahan17a}, our M22 has the same workflow structure, meaning there is one central server constantly communicating with several remote clients, while the DNN model architectures are the same in each client and the server. 
When initiating the FL training, the server randomly initializes the weights of the model. Afterwards, the training iteration begins with the server sending the global model to all the clients through the communication channel, which is known as down-link communication. 
Although starting with the same model parameters, the remote clients perform training on local data, and end with different parameters after local training.. Each client sends its distinct model back to the central server through the communication channel, which is known as uplink communication. Once all the updates are received, the central sever computes a weighted sum of the local model parameters and updates the global model, which indicates the completion of one FL training iteration. The next iteration begins with the server down-link transmitting the global model and ends up with the server updating the global model after the clients' uplink transmission of local models.

Our M22 targets on reducing the communication overhead between the server and remote clients, particularly in the uplink transmission. Leveraging the computation capability of the edge device, M22 approach guides how the remote clients compress the models before transmitting, and how the server recovers the compressed local models, which is detailed shown in the Algorithm \ref{alg:cap}.

\newlength\myindent
\setlength\myindent{2em}
\newcommand\bindent{%
  \begingroup
  \setlength{\itemindent}{\myindent}
  \addtolength{\algorithmicindent}{\myindent}
}
\newcommand\eindent{\endgroup}

\setlength\myindent{2em}

\begin{algorithm}
\caption{M22. \ The $\emph K$ clients are indexed by $\emph k$; $\emph B$ is the local mini-batch size; $\dv$  is the local dataset of each client; $\emph E$ is the number of local epochs; And $\eta$ is the learning rate.}\label{alg:cap}
\begin{algorithmic}
    \STATE \textbf{Server executes:}
    \bindent
    \STATE {initialize $\omega_0$}
    \FOR{each round $\emph t = 1,2, ...$}
    \FOR{each client $\emph k \in K$ in parallel}
    \STATE $\hat{g}^{k}_{t} \leftarrow \textrm{comp}_R^{-1}(\textrm{comp}_R (g^{k}_{t}))$
    \ENDFOR
    \STATE $\hat{\omega}_{t+1} \leftarrow \hat{\omega}_{t} - \frac{1}{n} \, \sum_{k=1}^{K} \hat{g}^{k}_{t}$
    \ENDFOR
    \eindent

    \STATE
    \STATE \textbf{Client executes:} \emph{// run on each client k}
    \bindent
    \STATE $\omega_{t} \leftarrow  \textrm{download from server}$ 
    \STATE $\omega_{t, sparse} \leftarrow topK(\omega_{t})$ topK sparsification
    \FOR{local iteration $e = 1$ to $E$}
    \FOR{each batch $b \in B$}
    \STATE $g^{k}_{t} \leftarrow \eta \,  \Lcal(\dv_{b}, \omega_{t, sparse})$  local training
    \ENDFOR
    \ENDFOR
    \FOR{each layer $g^{k}_{t,l}$ in $g^{k}_{t}$}
    \STATE fitting distribution + kmeans quantization
    \STATE distribution parameter $\leftarrow fitDistribution(g^{k}_{t,l})$
    \STATE centers,thresholds $\leftarrow k-means(\textrm{distribution parameter})$
    \STATE $\textrm{comp}_R (g^{k}_{t,l}) \leftarrow quantization(\textrm{centers, thresholds})$
    \ENDFOR
    \STATE transmit $\textrm{comp}_R (g^{k}_{t})$ to server
    \eindent
\end{algorithmic}
\end{algorithm}

\subsection{Further Comments}

In the workflow pipeline of M22, various elements could affect the performance. The hyper-parameters of FL setting, including the number of remote clients, the mini-batch size, the number of local training epoch and the model optimizer choice. As described in \ref{sec:DNN training}, we chose a naive setting for the major part of our simulations, merely to show the effectiveness of M22. It has been tested that M22 could be adapted in more complicated FL settings, including where partial clients are selected in each round, multiple training epochs are performed locally and the local datasets are heterogeneous. The choice of M is also considered as a hyper-parameter of the compression setting. 

Inspired by the \cite{SGD_sparse2}, where memory is equipped with SGD under compression, we implemented memory with M22. Each local client keeps a copy of the difference between its trained local model and the compressed version. This client-wise distinct difference is added back to downloaded global model by each client before the local training. The problems with this mechanism in the FL setting are memory accumulation and local optima. The memory stored by the clients would be accumulated throughout the training process, which could cause memory explosion. Additionally, the clients could converge to local optima in different directions. The server aggregation serves helps rectify the progressing directions to a global one, while an uncalibrated memory implementation would cause each client heading to different directions again. In our simulations, a fine-tuned memory weights could help the model convergence but much less significantly comparing with tuning the M value.


\section{{Numerical Evaluations}}
\label{sec:Numerical Evaluations}

%
%

%
%
%
In this section, we compare the performance of our proposed M22 algorithm with other ML compression techniques including (I) conventional floating point conversion, (i) $\topK$ sparsification, (ii)  sketching,  and (ii) non-uniform quantization using the $L_2$ norm.

To ensure the fairness of the comparison, the algorithms are implemented under the same FL setting and communication overhead constraint.
The training parameters are kept consistently across numerical experiments.

\subsection{{Compression Strategies Benchmark}}
\label{sec:Compression strategies}
In our simulations, we consider the following gradient compression techniques for benchmarking. 

\smallskip
\noindent
$\bullet$
\emph{$\topK$ sparsification + floating point (fp) representation:} In the context of \eqref{eq:comp}, $\topK$ can be applied to meet the rate constraint only once a certain format for representing the gradient entries has been established. Therefore, we consider a fp representation of the entries with $8$ and $4$ bits. Accordingly, the relationship between the sparsification parameter $K$, the digit precision $p$ (in bits), and the rate constraint $\Rsf$ in \eqref{eq:G}  is
\ea{
d \Rsf = \log \binom{d}{K_{\rm fp}} + K_{\rm fp} p.\label{eq:topk2}
}

\noindent
$\bullet$
\emph{$\topK$ sparsification  + scalar uniform quantization:} For the uniform quantizer with a given quantizer rate $R_{\rm u}$, the $2^{R_{\rm u}}$ quantization centers are uniformly distributed between the minimum and maximum values of the samples in each layer and each iteration.
The sparsification level $K_{\rm u}$ is accordingly chosen such that 
\ea{
d \Rsf = \log \binom{d}{K_{\rm u}} + K_{\rm u} R_{\rm u}. \label{eq:topk1}
}

\noindent
$\bullet$
\emph{Count Sketch }-- \cite{NEURIPS2019_75da5036}:
%
%
%
Inspired by \cite{NEURIPS2019_75da5036}, where the count sketch method is associated with SGD in the distributed learning scenario, we integrate it into our gradient compression workflow to make another comparison to our M22 algorithm. Following a $\topK$ sparsification of level $K_{\rm sk}$, each client performs local compression to its local stochastic gradient by count sketch via a common sketching operator; and the server recovers the indices and the values of large entries of the aggregated stochastic gradient from the gradient sketches. By choosing sketching ratio $r_{\rm sk} \in (0,1]$ of the common count sketch operator, the communication overhead is calculated as the following:
\ea{
d \Rsf = \log \binom{d}{K_{\rm sk}} +  r_{\rm sk} K_{\rm sk}.\label{eq:topk3}
}

\noindent
$\bullet$
\emph{TINYSCRIPT -- }\cite{fu2020don}:
In \cite{fu2020don}, the authors introduced a non-uniform quantization algorithm, TINYSCRIPT to compress the activations and gradients of a DNN.
In gradient compression, this scheme does not consider sparsification and applies $K$-means clustering to each layer before scalar quantization. Finally, the scalar quantizer hat minimizes the $L_2$ loss is employed.
We note that the layer-wise clustering is rather computationally expensive and the execution time suffers dramatically. For this reason, in our benchmark, we removed the clustering step so that the execution time of all scheme is comparable. 
After removing the step to cluster the gradients and adapting the approach to our FL setting, the workflow of TINYSCRIPT is similar to our M22 approach, and its communication overhead calculation could be categorized into the M22 and its variants.
We wish to point out that we have tried to adapt the schemes in \cite{NEURIPS2019_75da5036} and  \cite{fu2020don} to the setting of Sec. \ref{sec:System Model} so as to yield a fair comparison.
To comprehend the mechanism of TINTSCRIPT, we refer the readers to the original schemes \cite{fu2020don}.

\smallskip
\noindent
$\bullet$
\emph{M22 variants:} 
M22 has various incarnation, depending on the choice of M in \eqref{eq:our distortion} and the choice of fitting distribution -- GenNorm or Weibull.
%
The workflow of our proposed M22  algorithm includes $\topK$ sparsification and fitting each layer of the gradients with a pre-defined distribution. Fitting with GenNorm or Weibull distribution, the quantization centers will be distributed in a non-uniform manner. Changing from one distribution to another produces variants of the M22 algorithm, as well as adapting different choices of M value when computing gradient distortion. Because all the variants share a similar compression strategy, the calculation of $d \Rsf$  for them is the same, which means, for our proposed quantizer with a given rate $R_{\rm mw}$, a number of $2^{R_{\rm mw}}$ quantization centers at each layer and iteration are found by the $K$-means algorithm described in Section~\ref{sec:Quantizer design}. We use a $\topK$ sparsification before our compressor where the sparsification level $K_{\rm mw}$ satisfies the following: 
\ea{
d \Rsf = \log \binom{d}{K_{\rm mw}} + K_{\rm mw} R_{\rm mw },\label{eq:topk3}
}
where $R_{\rm M22}$ is the quantizer rate and $K_{\rm M22}$ is the sparsification level. 
\subsection{Simulation Results}
\label{sec:simu}
The results we demonstrate will show the superiority of our M22  in the various aspects including: improving the global model convergence comprehensively comparing with scalar quantization methods, uniform quantization methods and other adaptive quantization algorithms; expediting the convergence by flexibly tune the M value and exhibiting universal effectiveness for different DNN model architectures.
\noindent

$\bullet$ 
{\emph{ M22  vs. all:}}
In Fig.~\ref{fig:acc_rate1}, we plot the accuracy vs. iteration number under communication constraints of  $d\Rsf=332\text{kbits}$ and $d\Rsf=996\text{kbits}$ for the CNN network introduced in Table~\ref{tab:model_layers}. 
Such communication constraints are equivalent to allow each non-zero gradient to be represented using 1 bit and 3 bits, respectively.
In the following figures, each curve represents one of the compression strategies are as detailed in Sec. \ref{sec:Compression strategies}. Note that each accuracy result is the average of 5 different initializations. To match the rate constraint $d\Rsf=332\text{kbits}$, we choose the following parameters:

 \begin{itemize}
     \item $\topK$  + uniform: $R_{\rm u}=1$, $K_{\rm u}=331724$,
     \item $\topK$ + 8fp : $p=8$, $K_{\rm fp}=41466$,
     \item $\topK$ + 4fp : $p=4$, $K_{\rm fp}=82931$,
     \item M22 +GenNorm: $M=2$, $R_{\rm mw}=1$, $K_{\rm M22}=331724$,
     \item M22 +GenNorm: $M=3$, $R_{\rm mw}=1$, $K_{\rm M22}=331724$,
     \item TINYSCRIPT: $M=0$, $R_{\rm mw}=1$, $K_{\rm M22}=331724$,
     \item M22 +Weibull: $M=4$, $R_{\rm mw}=1$, $K_{\rm M22}=331724$,
     \item Count sketch: $r_{\rm sk}=1$, $K_{\rm sk}=331724$.
 \end{itemize}
\noindent
For dR = 996k bits, we choose the following parameters:
 
 \begin{itemize}
     \item $\topK$  + uniform: $R_{\rm u}=3$, $K_{\rm u}=331724$,
     \item $\topK$ + 8fp : $p=8$, $K_{\rm fp}=124396$,
     \item $\topK$ + 4fp : $p=4$, $K_{\rm fp}=248793$,
     \item M22 +GenNorm: $M=2$, $R_{\rm mw}=3$, $K_{\rm M22}=331724$,
     \item M22 +GenNorm: $M=9$, $R_{\rm mw}=3$, $K_{\rm M22}=331724$,
     \item TINYSCRIPT: $M=0$, $R_{\rm mw}=3$, $K_{\rm M22}=331724$,
     \item M22 +Weibull: $M=7$, $R_{\rm mw}=3$, $K_{\rm M22}=331724$,
     \item Count sketch: $r_{\rm sk}=3$, $K_{\rm sk}=331724$.
 \end{itemize}

Each approach described above is represented by one curve of a particular color as shown in the figure legends. Note that we use the letter "G" or "W" to indicate either GenNorm or d-Weibull distribution is applied in gradients distribution approximation, and the number behind is the M value we use throughout the training.

\begin{figure}[t!]
  \centering

\definecolor{mycolor3}{rgb}{0.00000,0.49804,0.00000}%

\begin{tikzpicture}   
\begin{axis}[
    scale only axis, 
    height=11 cm,
    width=0.45\linewidth,
	xtick={1,2,3,4,5,6,7,8,9,10},
	ytick={0.1,0.2,0.3,0.4,0.5,0.6,0.7,0.8},
	grid=both,
	minor grid style={gray!25},
	major grid style={gray!25},
	legend style={at={(1,0)},anchor=south east},
	no marks]
\addplot[line width=1pt,solid,color=magenta] %
	table[x=rounds,y=uniform,col sep=comma]{./figures_data/aa_acc_rate1.csv};
\addplot[line width=1pt,solid,color=yellow] %
	table[x=rounds,y=fp8,col sep=comma]{./figures_data/aa_acc_rate1.csv};
\addplot[line width=1pt,solid,color=brown] %
	table[x=rounds,y=fp4,col sep=comma]{./figures_data/aa_acc_rate1.csv};
\addplot[line width=1pt,solid,color=black] %
	table[x=rounds,y=SMG_GenNormM2,col sep=comma]{./figures_data/aa_acc_rate1.csv};
\addplot[line width=1pt,solid,color=mycolor3] %
	table[x=rounds,y=SMG_GenNormM3,col sep=comma]{./figures_data/aa_acc_rate1.csv};
\addplot[line width=1pt,solid,color=blue] %
	table[x=rounds,y=sketch,col sep=comma]{./figures_data/aa_acc_rate1.csv};
\addplot[line width=1pt,solid,color=gray] %
	table[x=rounds,y=TINYSCRIPT,col sep=comma]{./figures_data/aa_acc_rate1.csv};
\addplot[line width=1pt,solid,color=red] %
	table[x=rounds,y=SMG_WeibM4,col sep=comma]{./figures_data/aa_acc_rate1.csv};
\addlegendentry{uniform};
\addlegendentry{$8$fp};
\addlegendentry{$4$fp};
\addlegendentry{M22-G2};
\addlegendentry{M22-G3};
\addlegendentry{count sketch};
\addlegendentry{TINYSCRIPT};
\addlegendentry{M22-W4};
\end{axis}
\end{tikzpicture}
\begin{tikzpicture}  
\begin{axis}[
    scale only axis, 
    height=11 cm,
    width=0.45\linewidth,
	xtick={1,2,3,4,5,6,7,8,9,10},
	ytick={0.1,0.2,0.3,0.4,0.5,0.6,0.7,0.8},
	grid=both,
	minor grid style={gray!25},
	major grid style={gray!25},
	legend style={at={(1,0)},anchor=south east},
	no marks]
\addplot[line width=1pt,solid,color=magenta] %
	table[x=rounds,y=uniform,col sep=comma]{./figures_data/aa_acc_rate3.csv};
\addplot[line width=1pt,solid,color=yellow] %
	table[x=rounds,y=fp8,col sep=comma]{./figures_data/aa_acc_rate3.csv};
\addplot[line width=1pt,solid,color=brown] %
	table[x=rounds,y=fp4,col sep=comma]{./figures_data/aa_acc_rate3.csv};
\addplot[line width=1pt,solid,color=black] %
	table[x=rounds,y=SMG_GenNormM2,col sep=comma]{./figures_data/aa_acc_rate3.csv};
\addplot[line width=1pt,solid,color=mycolor3] %
	table[x=rounds,y=SMG_GenNormM9,col sep=comma]{./figures_data/aa_acc_rate3.csv};
\addplot[line width=1pt,solid,color=blue] %
	table[x=rounds,y=sketch,col sep=comma]{./figures_data/aa_acc_rate3.csv};
\addplot[line width=1pt,solid,color=gray] %
	table[x=rounds,y=TINYSCRIPT,col sep=comma]{./figures_data/aa_acc_rate3.csv};
\addplot[line width=1pt,solid,color=red] %
	table[x=rounds,y=SMG_WeibM7,col sep=comma]{./figures_data/aa_acc_rate3.csv};
\addlegendentry{uniform};
\addlegendentry{$8$fp};
\addlegendentry{$4$fp};
\addlegendentry{M22-G2};
\addlegendentry{M22-G9};
\addlegendentry{count sketch};
\addlegendentry{TINYSCRIPT};
\addlegendentry{M22-W7};
\end{axis}
\end{tikzpicture} 
  \caption{Comparison  across different gradient compression approaches under rate constraint dR = 332k (left) bits and dR = 996k bits (right).}
 \label{fig:acc_rate1}
\end{figure}
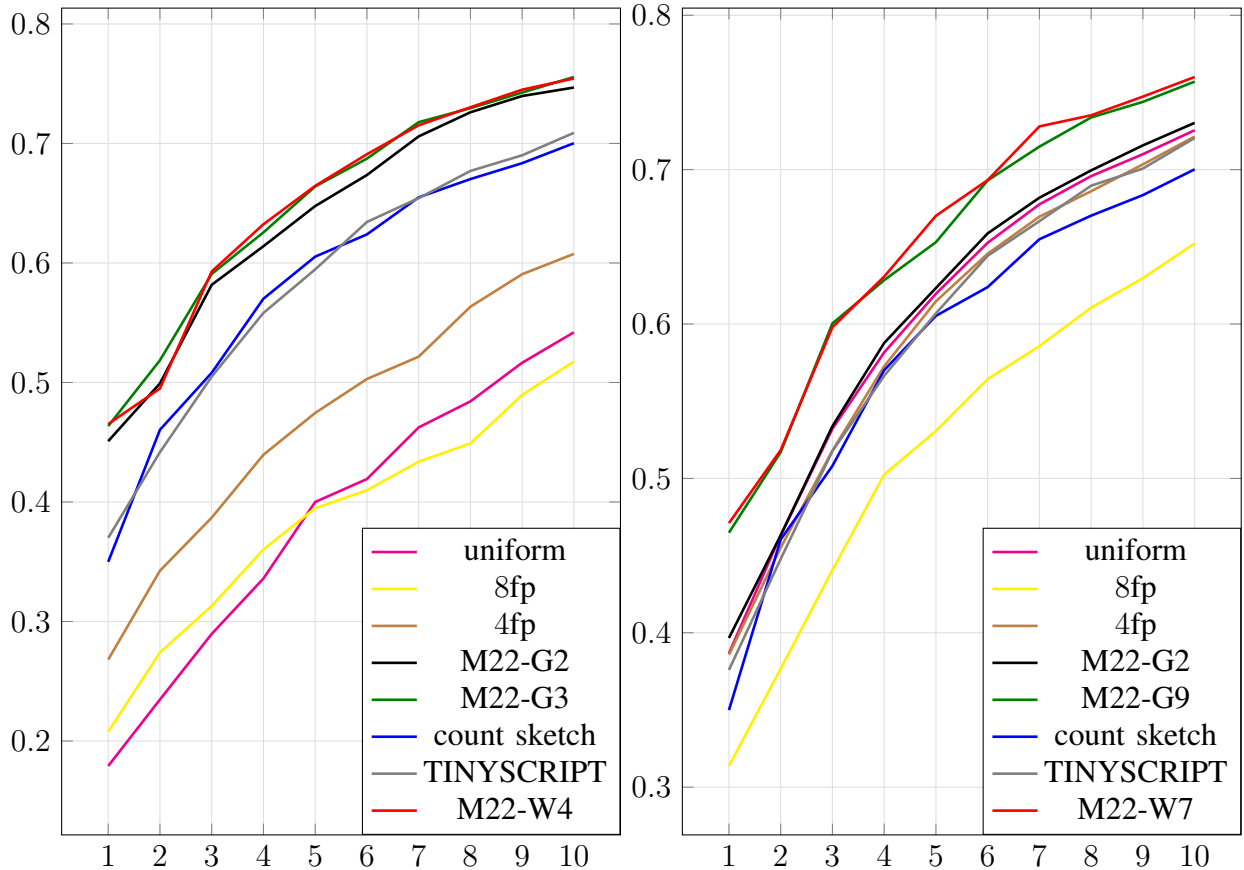

We point out that non-uniform quantization algorithms generally have better performance.  
In the left panel of Fig.~\ref{fig:acc_rate1}, when the communication overhead constraint is set to $d\Rsf=332\text{kbits}$ for CNN training, each nonzero gradient update is represented using just $1$ bit. Under such a restricted condition, where most of the traditional approaches converge reluctantly, our M22  algorithm with a fine-tuned $M$ value not only outperforms other approaches, but also maintains a similar converging speed comparing with a loose-constraint regime.
In the right panel of Fig.~\ref{fig:acc_rate1}, when the communication overhead constraint is relaxed to $d\Rsf=996\text{kbits}$, the performance of traditional and scalar approaches start to be competitive with count sketch and TINYSCRIPT. However, our M22 curves stay on top by an obvious margin comparing with all others.
Note that the quantizer for M22  and TINYSCRIPT is chosen adaptively according to \ref{sec:Quantizer design}, chosen as a function of the empirical gradient distribution at each iteration. 
For the case of M22  GenNorm, this is attained by pre-calculating the quantization centers for different values of shape parameter $\be$ of GenNorm distribution \eqref{eq:gennorm}.
At each iteration, the gradient vector is normalized to obtain a zero-mean unit-variance vector which is then quantized using the pre-calculated quantizer.

\noindent
$\bullet$
 {\emph {M's effect:}} After conducting numerous simulations, the M value \eqref{eq:our distortion} has been verified to play a vital role in model convergence speed and final accuracy. Recall that we enable the flexibility of choosing the M value when designing the gradients distortion measurement in \ref{sec:Distortion Measure}. In Fig.~\ref{fig:acc_rate1}, there are two curves of M22  using GenNorm distribution with two different M values. Increasing the M value would result in more sparse quantization centers and thresholds, consistent with the fact that gradients are long-tailed, and therefore achieve better performance. The same effect could be found when comparing TINYSCRIPT with M22  using d-Weibull distribution. 
 TINYSCRIPT approach can be regarded as a degenerate version of our M22 with $\textrm{M}=0$ when calculating the distortion, while a fined-tuned positive M value is applied in our M22, which always yields a better result. 
 
 In Fig.~\ref{fig:Meffect}, we employ a list of M choices to demonstrate how the value impact the convergence substantially. 
 Fitting the gradients with GenNorm distribution and limiting the communication overhead to $d\Rsf=664\text{kbits}$, the performance of $\textrm{M}=6$ is the best in terms of final accuracy. However, in the first a few rounds, $\textrm{M}=8$ achieves the highest accuracy. According to our extensive simulation results, large values of $M$ could boost the model convergence in the initial training phase. However, keeping it too large would harm the model convergence after certain stages. By adapting our M22  algorithm, a fine-tuned $M$ value would not only result in fast improvement at the early training process, but also ensure the final accuracy. The same effect of M value could be found in the cases where the Weibull distribution is adapted to approximate the gradients distribution and using other model structures. 

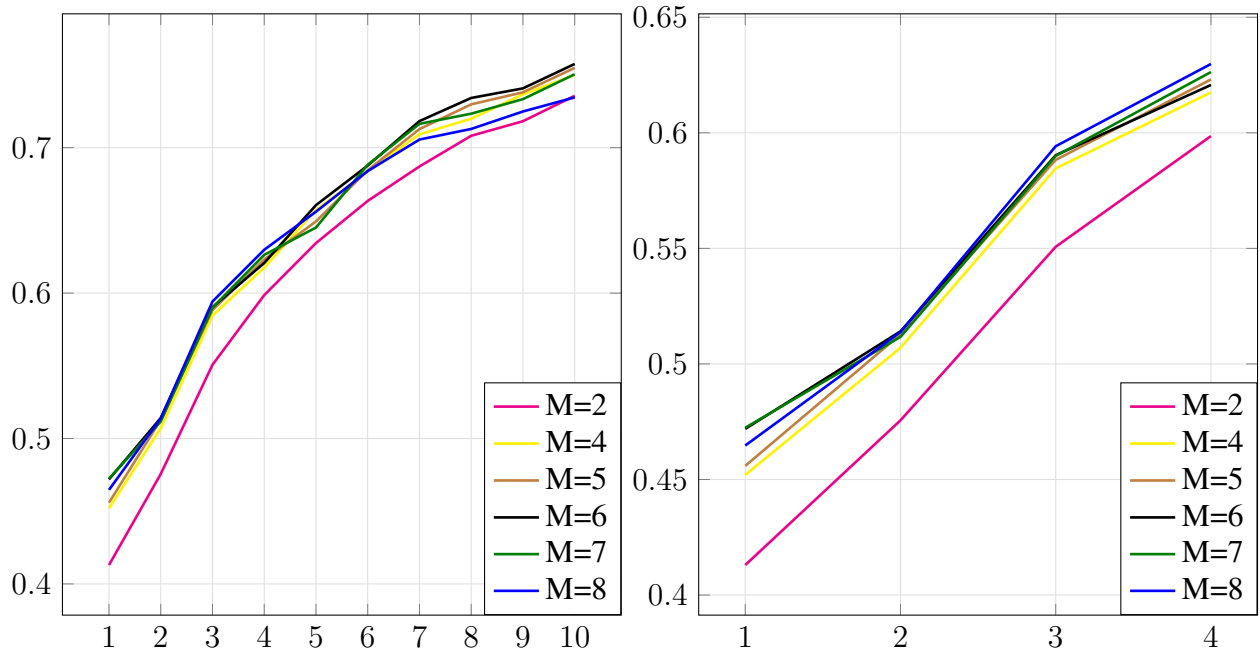
\begin{figure}[t!]
  \centering

\definecolor{mycolor3}{rgb}{0.00000,0.49804,0.00000}%

\begin{tikzpicture}   
\begin{axis}[
    scale only axis, 
    height=8cm,
    width=0.45\linewidth,
	xtick={1,2,3,4,5,6,7,8,9,10},
	ytick={0.1,0.2,0.3,0.4,0.5,0.6,0.7,0.8},
	grid=both,
	minor grid style={gray!25},
	major grid style={gray!25},
	legend style={at={(1,0)},anchor=south east},
	no marks]
\addplot[line width=1pt,solid,color=magenta] %
	table[x=rounds,y=M2,col sep=comma]{./figures_data/aa_Meffect_10.csv};
\addplot[line width=1pt,solid,color=yellow] %
	table[x=rounds,y=M4,col sep=comma]{./figures_data/aa_Meffect_10.csv};
\addplot[line width=1pt,solid,color=brown] %
	table[x=rounds,y=M5,col sep=comma]{./figures_data/aa_Meffect_10.csv};
\addplot[line width=1pt,solid,color=black] %
	table[x=rounds,y=M6,col sep=comma]{./figures_data/aa_Meffect_10.csv};
\addplot[line width=1pt,solid,color=mycolor3] %
	table[x=rounds,y=M7,col sep=comma]{./figures_data/aa_Meffect_10.csv};
\addplot[line width=1pt,solid,color=blue] %
	table[x=rounds,y=M8,col sep=comma]{./figures_data/aa_Meffect_10.csv};
\addlegendentry{M=2};
\addlegendentry{M=4};
\addlegendentry{M=5};
\addlegendentry{M=6};
\addlegendentry{M=7};
\addlegendentry{M=8};
\end{axis}
\end{tikzpicture}
\begin{tikzpicture}  
\begin{axis}[
    scale only axis, 
    height=8cm,
    width=0.45\linewidth,
	xtick={1,2,3,4},
	grid=both,
	minor grid style={gray!25},
	major grid style={gray!25},
	legend style={at={(1,0)},anchor=south east},
	no marks]
\addplot[line width=1pt,solid,color=magenta] %
	table[x=rounds,y=M2,col sep=comma]{./figures_data/aa_Meffect_4.csv};
\addplot[line width=1pt,solid,color=yellow] %
	table[x=rounds,y=M4,col sep=comma]{./figures_data/aa_Meffect_4.csv};
\addplot[line width=1pt,solid,color=brown] %
	table[x=rounds,y=M5,col sep=comma]{./figures_data/aa_Meffect_4.csv};
\addplot[line width=1pt,solid,color=black] %
	table[x=rounds,y=M6,col sep=comma]{./figures_data/aa_Meffect_4.csv};
\addplot[line width=1pt,solid,color=mycolor3] %
	table[x=rounds,y=M7,col sep=comma]{./figures_data/aa_Meffect_4.csv};
\addplot[line width=1pt,solid,color=blue] %
	table[x=rounds,y=M8,col sep=comma]{./figures_data/aa_Meffect_4.csv};
\addlegendentry{M=2};
\addlegendentry{M=4};
\addlegendentry{M=5};
\addlegendentry{M=6};
\addlegendentry{M=7};
\addlegendentry{M=8};
\end{axis}
\end{tikzpicture} 
  \caption{10-round accuracy comparison across different $M$ values (left) and zoom-in first 4-round accuracy (right) under the communication constraint $d\mathsf{R}=664\text{kbits}$.
  }
 \label{fig:Meffect}
\end{figure}

\noindent
$\bullet$
 {\emph {ResNet18 \& VGG16:}} Besides the CNN network, our M22  algorithm guarantees fast and reliable global model convergence when applied to other conventional network architectures, including ResNet and VGG networks. In the left panel of Fig.~\ref{fig:resvgg}, under the same communication constraints, we compare the 3 non-uniform compression algorithms: count sketch, TINYSCRIPT and M22 using ResNet18 model. Speaking of the converging speed and accuracy, our M22  algorithm adapting with GenNorm distribution fitting performs very close to TINYSCRIPT, which estimates the gradient updates with a Weibull distribution, while count sketch approach falls behind by an obvious margin. In the right panel of Fig.~\ref{fig:resvgg}, we compare the scheme where no quantization is applied with our M22  approach under 4 different communication overhead constraints, including $d\Rsf=332\text{kbits}$, $d\Rsf=664\text{kbits}$, $d\Rsf=996\text{kbits}$ and $d\Rsf=1.33\text{mbits}$.
 On one hand, when the communication constraint is loose, M22  algorithm performs almost as good as the case, where there is no quantization applied. On the other hand, with one-eighth of the communication overhead, although the convergence process inevitably becomes less smooth, our M22  is still competitive in terms of progressing speed, and it does not take too long until it achieves similar final accuracy comparing to more relaxed-constraint cases.
 

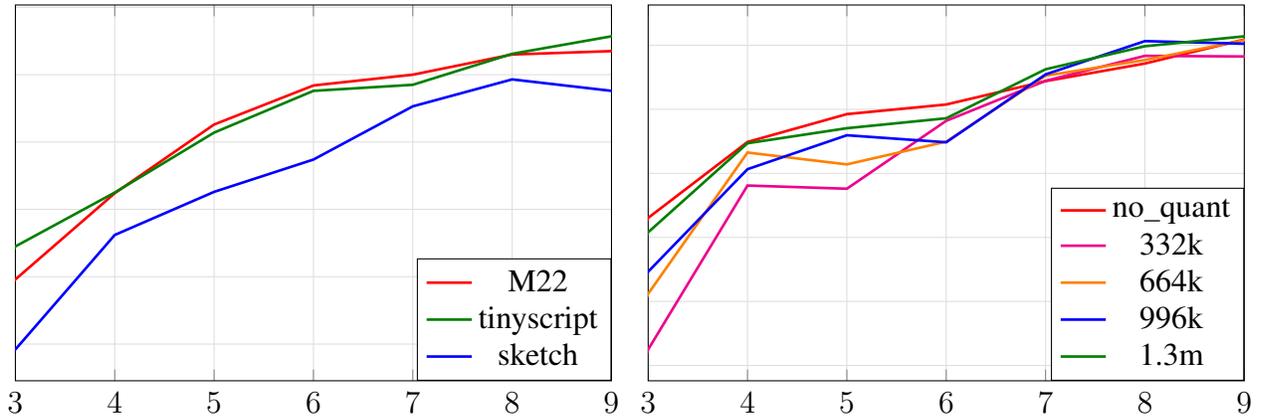
\begin{figure}[t!]
  \centering
 \definecolor{mycolor3}{rgb}{0.00000,0.49804,0.00000}%

\begin{tikzpicture}   
\begin{axis}[
    scale only axis, 
    height=5cm,
    width=0.48\linewidth,
	xtick={3,4,5,6,7,8,9},
	xmax=9,
    xmin=3,
	ytick={},
	ymajorticks=false,
	grid=both,
	minor grid style={gray!25},
	major grid style={gray!25},
	legend style={at={(1,0)},anchor=south east},
	no marks]
\addplot[line width=1pt,solid,color=red] %
	table[x=rounds,y=SMG,col sep=comma]{./figures_data/aa_resnet18.csv};
\addplot[line width=1pt,solid,color=mycolor3] %
	table[x=rounds,y=tinyscript,col sep=comma]{./figures_data/aa_resnet18.csv};
\addplot[line width=1pt,solid,color=blue] %
	table[x=rounds,y=sketch,col sep=comma]{./figures_data/aa_resnet18.csv};
\addlegendentry{M22};
\addlegendentry{tinyscript};
\addlegendentry{sketch};
\end{axis}
\end{tikzpicture}
\begin{tikzpicture}  
\begin{axis}[
    scale only axis, 
    height=5cm,
    width=0.48\linewidth,
	xtick={3,4,5,6,7,8,9},
	xmax=9,
    xmin=3,
	ytick={},
	ymajorticks=false,
	grid=both,
	minor grid style={gray!25},
	major grid style={gray!25},
	legend style={at={(1,0)},anchor=south east},
	no marks]
\addplot[line width=1pt,solid,color=red] %
	table[x=rounds,y=no_quant,col sep=comma]{./figures_data/aa_vgg16.csv};
\addplot[line width=1pt,solid,color=magenta] %
	table[x=rounds,y=r1,col sep=comma]{./figures_data/aa_vgg16.csv};
\addplot[line width=1pt,solid,color=orange] %
	table[x=rounds,y=r2,col sep=comma]{./figures_data/aa_vgg16.csv};
\addplot[line width=1pt,solid,color=blue] %
	table[x=rounds,y=r3,col sep=comma]{./figures_data/aa_vgg16.csv};
\addplot[line width=1pt,solid,color=mycolor3] %
	table[x=rounds,y=r4,col sep=comma]{./figures_data/aa_vgg16.csv};
\addlegendentry{no\_quant};
\addlegendentry{332k};
\addlegendentry{664k};
\addlegendentry{996k};
\addlegendentry{1.3m};
\end{axis}
\end{tikzpicture} 
 \vspace{-0.25cm}
  \caption{Compare the 3 non-uniform compression algorithms on ResNet18 (left); Compare the scheme where no quantization is applied with our M22  approach under different communication overhead constraints on VGG16 (right).}
  
 \label{fig:resvgg}
\end{figure}

 \section{Conclusion}
\label{sec:conclusion}

In this paper, the problem of efficient gradient compression for federated learning has been considered. 
For this problem, we propose M22 as an efficient gradient compression algorithm developed from a rate-distortion perspective.
More specifically, M22 tackles gradient compression by designing a quantizer under (i) an assumption on the distribution of the model gradient updates and (ii) a choice of the distortion measure which minimizes the loss in accuracy. 
Regarding the gradients distribution, we assume that the gradient updates in each model layer and at each iteration follow a 2-degree of freedom distribution, either generalized normal distribution or 
double-sided Weibull distribution.
This assumption is verified by numerous simulation results we produced, and therefore makes us confident to conclude that the extra degree of freedom is essential for approximating the model gradients, especially under the federated learning setting.
In terms of the distortion measure, we assume that the distortion capturing the relationship between gradient perturbation and loss in accuracy is the $M$-magnitude weighted $L_2$ distortion, that is the $L_2$ loss between the compressed and original gradient updates multiplied by the magnitude of the original gradient to the power of $M$. 
We argue that this choice of distortion naturally bridges between two classical gradient sparsification approaches: $M=0$ recovers uniform quantization, while $M \goes \infty$ recovers $\topK$ sparsification. 
Our simulations show that the choice of $M$ plays an important role in the federated learning scheme, where small values are preferred when we need to compress the models aggressively, while larger values are more appropriate under loose communication overhead constraints.
In this work, both of these assumptions are validated through numerical evaluations under different federated learning settings and different model structures. The performance of M22 is competitive comparing with other state-of-the-art methods in most cases and surpasses them in a few. 
Simulations under more complicated federated learning settings definitely require more computation power. Finally, a more theoretical justification of these two assumptions will be investigated in our future research.

\newpage
\bibliographystyle{IEEEtran}
\bibliography{FL_lattice,new_bib}



\end{document}